\documentclass[lettersize,journal]{IEEEtran}
\usepackage{amsmath,amsfonts}
\usepackage{algorithmic}
\usepackage{algorithm}
\usepackage{array}
\usepackage[caption=false,font=normalsize,labelfont=sf,textfont=sf]{subfig}
\usepackage{textcomp}
\usepackage{stfloats}
\usepackage{url}
\usepackage{verbatim}
\usepackage{graphicx}
\usepackage{yhmath}
\usepackage{cite}
\usepackage{booktabs}
\usepackage{multirow}

\makeatletter
\let\NAT@parse\undefined
\makeatother
\usepackage{hyperref}
\usepackage{amssymb}
\begin{document}

\title{A Novel Generative Convolutional Neural Network for Robot Grasp Detection on Gaussian Guidance}

\author{Yuanhao Li,~Yu Liu,~Zhiqiang Ma, \emph{Member, IEEE},~Panfeng Huang, \emph{Senior Member, IEEE}
	\thanks{
	
)

}
}

\maketitle

\begin{abstract}
The vision-based grasp detection method is an important research direction in the field of robotics. However, due to the rectangle metric of the grasp detection rectangle's limitation, a false-positive grasp occurs, resulting in the failure of the real-world robot grasp task. In this paper, we propose a novel generative convolutional neural network model to improve the accuracy and robustness of robot grasp detection in real-world scenes. First, a Gaussian-based guided training method is used to encode the quality of the grasp point and grasp angle in the grasp pose, highlighting the highest-quality grasp point position and grasp angle and reducing the generation of false-positive grasps. Simultaneously, deformable convolution is used to obtain the shape features of the object in order to guide the subsequent network to the position. Furthermore, a global-local feature fusion method is introduced in order to efficiently obtain finer features during the feature reconstruction stage, allowing the network to focus on the features of the grasped objects. On the Cornell Grasping Datasets and Jacquard Datasets, our method achieves excellent performance of 99.0$\%$ and 95.9$\%$, respectively. Finally, the proposed method is put to the test in a real-world robot grasping scenario.
\end{abstract}


\section{Introduction}\label{sec1}
\IEEEPARstart{W}{ith} the development of information technology and control technology, robots are becoming more important in industrial manufacturing~\cite{1}, medical assistance~\cite{2}, social service~\cite{3}, and space exploration~\cite{4}. Grasping objects is the most widely used robot task and one of the most challenging technologies in robot operation. Although the action of ``picking up'' and ``putting down'' is a very basic behavior for humans, for robots it involves a series of detection, planning, and control systems. To complete the grasping task, the robot must first perceive the object, just as a human does with his eyes, in order to determine important information such as its position, direction, and grasping position. As a result, the grasp detection system serves as a starting point from which to plan future grasping paths and carry out the whole grasping action.

\begin{figure}[!t]
	\centering
	\includegraphics[width=6cm]{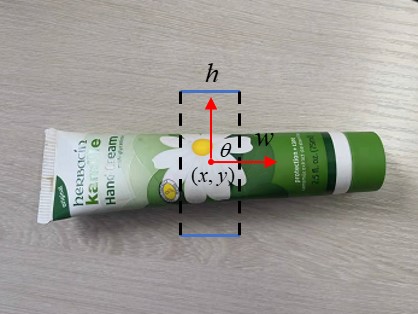}
	\caption{The oriented rectangle representation of grasp. where $(x, y)$ represent the center position of the grasp, $(h, w)$ represent the size and opening length of the gripper, and $\theta$ represents the intersection angle between the direction of the robot parallel plate gripper expansion and the horizontal axis.}\label{fig1}
\end{figure}

The key to the success of the grasping operation is obtaining an accurate grasping position. The early research on grasp detection is based on the analysis method~\cite{5}, which is based on the analysis of geometric and physical characteristics as well as manual design features to achieve the best grasp point selection. Although grasping accuracy is improved for specific objects, the manually-designed features are cumbersome and time-consuming, with poor generalization ability and weak universality. Deep learning development provides a new way to solve generalization problems and has made significant progress in many fields, such as object detection~\cite{6,7,8}, demonstrating a strong ability for feature extraction. Object detection methods, on the other hand, typically only output a unique ground truth to express the detection position and category of objects. However, this does not apply to the grasped object because an object does not usually have only one grasping position. As a result, how the grasping representation of objects is designed can effectively deal with the robot grasping task in an unstructured environment. In 2014, Lenz et al.~\cite{9} proposed a rectangle box description of the grasping position with reference to Jiang et al.~\cite{10}, as shown in Fig.~\ref{fig1}, to transform grasping detection into a problem similar to object detection. Lenz first found all possible grasp rectangles using a shallow convolutional neural network, and then used a deep convolutional neural network to find a grasp rectangle that is in line with the rectangle metric and thus a good grasp rectangle. The rectangle metric is as follows: The angle difference between the predicted and ground truth grasp angles is less than $30^{\circ}$; 2) The Jaccard index of the predicted and ground truth grasps is greater than 25\%; the Jaccard index is defined as:
\begin{equation}\label{equ1}
Jac=\frac{{area}(A \cap B)}{{area}(A \cup B)}
\end{equation}
where $A$, $B$ represent the predicted and ground truth respectively. Most of the subsequent grasp detection methods are developed based on the five-dimensional grasp representation proposed by Lenz and adopt the rectangle metric. 

However, we note the limitations of the rectangle metric. In fact, whether it is ``greater than 0.25'' or ``less than $30^{\circ}$,'' there is a loose threshold, which may lead to false-positive grasps. Although the prediction rectangle meets the rectangle metric standard, as shown in Fig. 2a, the center point of the prediction rectangle is far from the center point of ground truth, and the center point of the prediction rectangle is not on the object. In a real-world robot grasping task, the gripper will come into contact with the object, which can cause the robot to fail to grasp and even cause damage to both the robot and the object. Similarly, as shown in Fig. 2b, even though the direction angle difference between the predicted rectangle's grasping center point and the ground truth's grasping center point is less than $30^{\circ}$, it is still an unreasonable grasping position. When humans pick up a regular object, they frequently choose to grasp at an angle perpendicular to the direction of the object, which is more stable than oblique grasping. Likewise, the gripper lacks the tactile sense of human fingers. The strength of grasping objects can be accurately assessed by human fingers. As a safety measure, the strength of the gripper is usually cut back. This requires a very precise and stable grasping position, or the object will fall while it is being grasped. Indeed, we have demonstrated this on the robot in previous work~\cite{11,12}. Although the majority of grasping research has achieved high performance on the Cornell Grasp Dateset and other datasets, grasping's actual success rate is extremely low. Along with the interference of some external factors (light conditions, ambient temperature, camera coordinate conversion, etc. ), a significant portion of the reason is due to the rectangle metric's limitations and the absence of detailed analysis of the entire object and grasping area. As a result, while we strive for high accuracy, we prefer to concentrate our efforts on increasing the grasping success rates in the real world.

\begin{figure}[!t]
	\centering
	\subfloat[]{\includegraphics[width=0.5\linewidth]{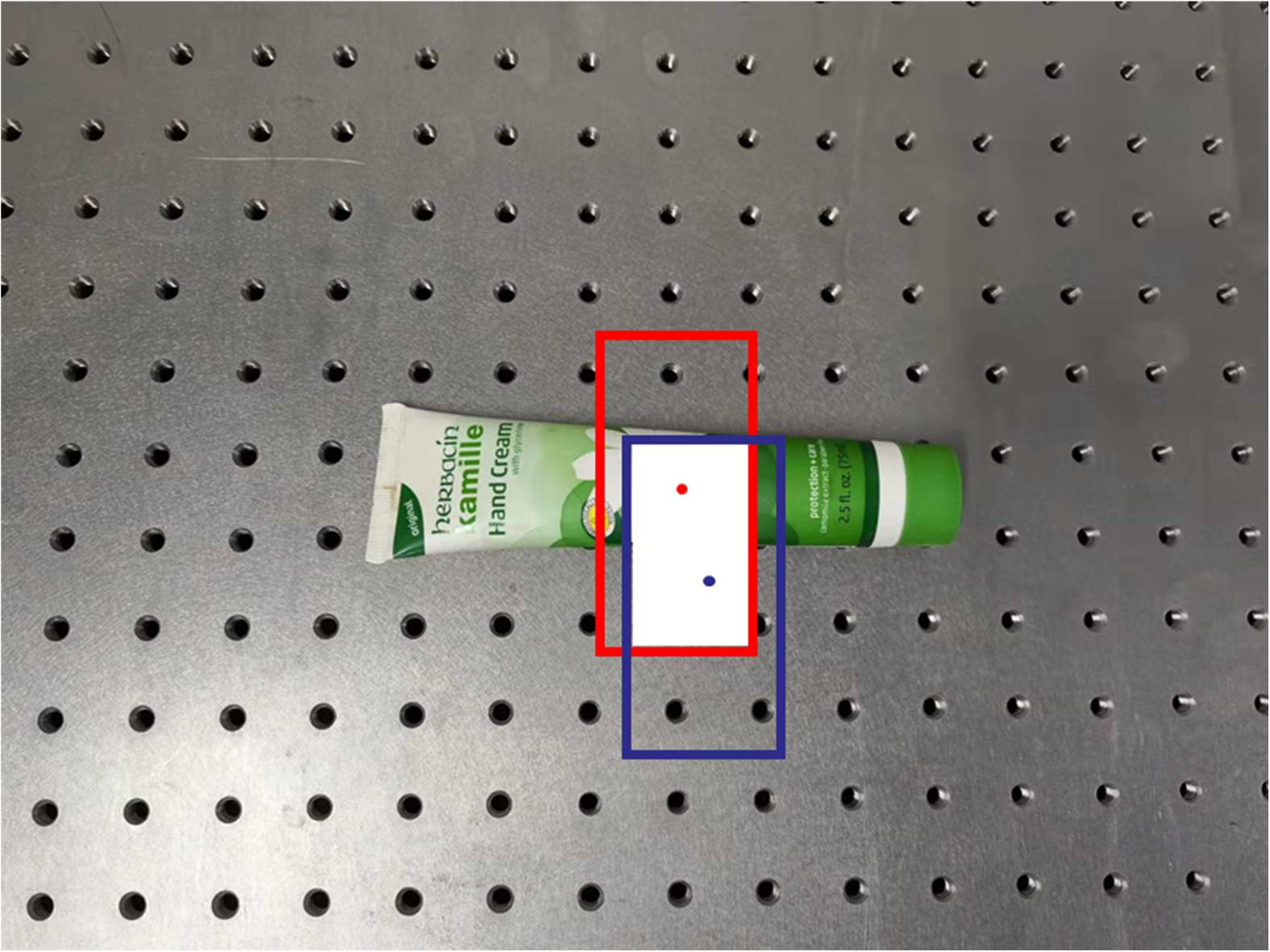}}
	\subfloat[]{\includegraphics[width=0.5\linewidth]{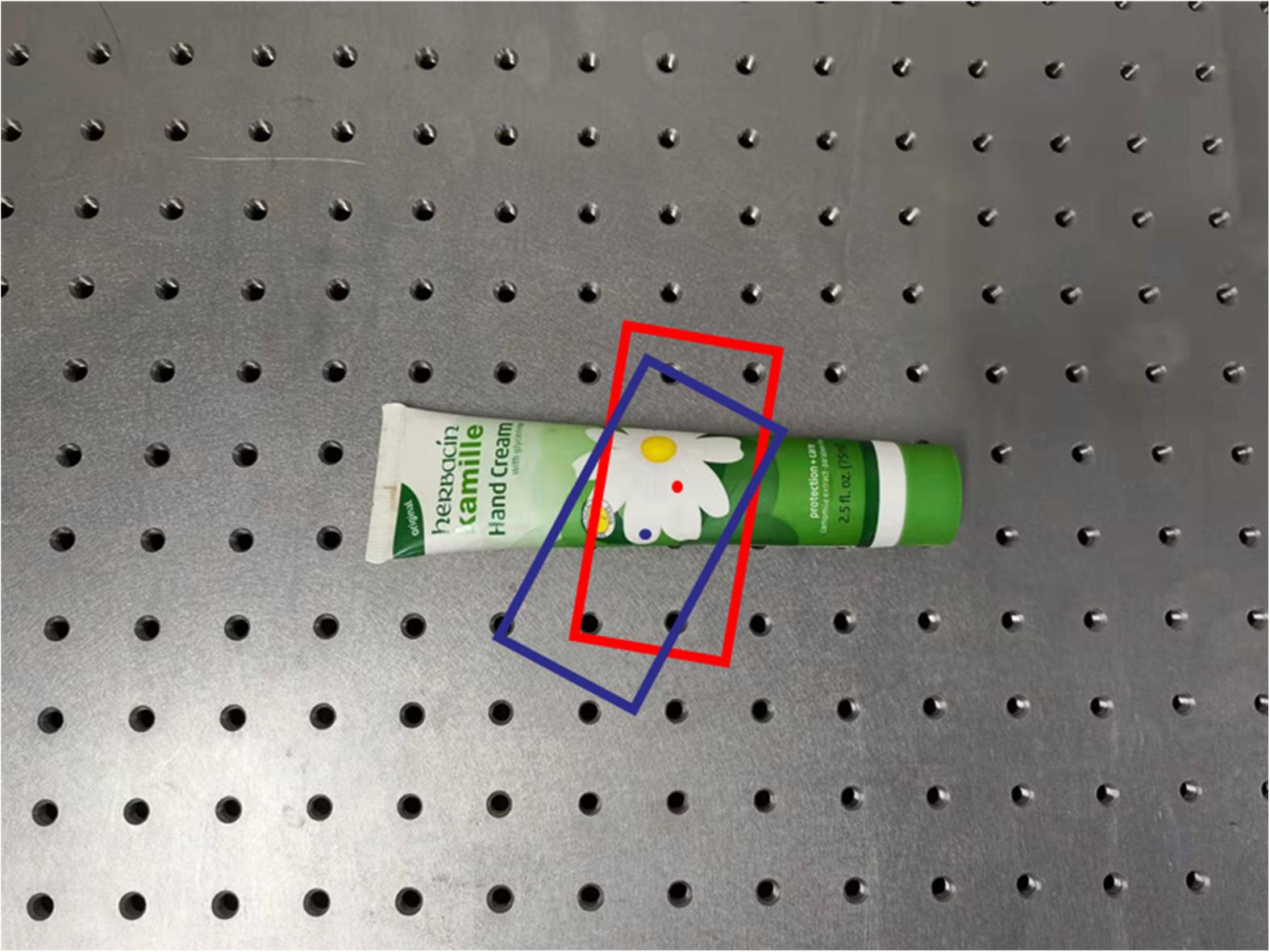}}
	\caption{Some examples of false-positive grasps. (a). The white intersection region between the predicted rectangle and ground truth accounts 33$\%$ of their union, but obviously the predicted rectangle is a example of failure. (b) The angle difference between the predicted and ground truth grasp angles is less than $30^{\circ}$, but in the actual experiment, it is a shaky grasp.} \label{fig2}
\end{figure}

In order to solve the above problems, a Gaussian-based guided training method is used to encode the quality of the grasp points and angles in a grasp pose, which directs the network's learning to focus on improving the quality of grasp points and angles for points on the object's center line and for points that are perpendicular to the object's center line. On this basis, a generative convolutional neural network model for grasping detection of real-scene robots is proposed. Simultaneously, a deformable convolution is introduced to obtain the object's shape features, which will be used to guide the subsequent network back to its original position. Additionally, a global-local feature fusion method is introduced to efficiently obtain finer features in the feature reconstruction stage, allowing the network to focus on the features of grasped objects. Finally, we propose a multi-scale loss function for grasp detection that predicts grasp positions at various scales in order to adapt to grasping objects of various scales in the real world.

In summary, the main contributions of this paper are as follows:
\begin{enumerate}
	\item  We use a Gaussian-based guided training mehod to improve the existing grasping representation, which standardizes the position and angle information of the grasping rectangle to the maximum extent possible and significantly improves the grasping success rate in real-world tests.
	\item To obtain refined object features, an attention network for global-local feature fusion is proposed, which divides feature enhancement into global and local branches.
	\item On the Cornell Grasp Dataset and the Jacquard Dataset, our method achieves excellent results, with 99.0$\%$ and 95.9$\%$, respectively. Simultaneously, the superiority of our method is demonstrated on a real robot grasp system.
\end{enumerate}

\section{Related Work}\label{sec2}
The research on object grasping position detection began in the 1980s, and most of the early researches were mainly focused on the detection of grasping points, which was using heuristic algorithms to grasp and detect objects with specific shapes~\cite{13,14}. These methods could only achieve good results for objects with shape characteristics, and they were not able to provide a definite description of the grasping method, and the generalization performance was poor. Jiang et al.~\cite{10} proposed a rectangle box description of grasping position, which transformed the problem of object grasping position detection into the problem of finding a rectangle in image space. However, this model required to design visual features manually for specific objects, instead of extracting the features of the grasped region in a self-learning method.

Deep learning method has been shown to be effective for a wide range of perceptual problems~\cite{15,16}, which allows the perceptual system to learn mappings from some feature sets to various visual features. More researchers are gradually introducing deep learning into grasp detection, applying learning methods to vision and introducing learning methods to score for grasp quality. Lenz et al.~\cite{9} first introduced a neural network as a classifier in the sliding window detection framework to predict whether a small block of image in the input image has a suitable grasping position. However, such a screening process could lead to fairly poor real-time performance, and the adopted network structure could also be more complicated. Redmon et al.~\cite{17} abandoned the framework based on the sliding window box and used the powerful feature extraction capability of the AlexNet to transform the prediction problem of the grasp region parameters into a regression problem, but this method could only predict a single grasp region for the input image, and its mapping mechanism often led to the average effect of grasping prediction results. In~\cite{18}, a shared convolutional neural network model was proposed to simultaneously complete the detection and classification tasks of grasped region, and the results showed that the performance of the parameter shared network was due to the single detection network. Chu et al.~\cite{19} used the “Grasp Region Proposal Network” to predict the undirected grasp candidate regions, and then delineated the rotation angle corresponding to the grasp candidate region from the perspective of classification, which could predict the grasp candidate box of multiple objects at the same time. Lan et al.~\cite{20} based a fully convolutional grasp part detection network on the directional anchor box, which would be more suitable for the detection of multi-angle grasping parts by adding angle information to the preset anchor box. On the basis of lenz, reference~\cite{21} directly generated pixel-level representations of grasping parameters through grasping generative convolutional neural network, and proposed a new type of representation method for grasp detection, thereby avoided the sampling process of grasp candidate regions which improved the detection efficiency significantly, and achieved a detection rate of 73\% in the Cornell dataset with only depth data as input. Reference~\cite{22} took RGB-D as input, and proposed a novel Generative Residual Convolutional Neural Network Model (GR-ConvNet) based on~\cite{21}, which achieved a detection rate of 97.7\% on the Cornell dataset. Chen et al.~\cite{12} proposed a two-stage grasp detection method, which only takes RGB images as input and utilizes low-level features and grasping criteria to select a small number of grasp candidates, and introduced a lightweight CNN model to evaluate grasp quality. Apart from that, tactile information has also been shown to play an important role in improving grasping success rate~\cite{23}.

Although the above methods have achieved good performance in public datasets, they lack detailed analysis of the overall object and grasp region, and the limitations of the rectangle metric will lead to excessive false-positive grasps, resulting in actual grasping failures. In the meantime, the focus of grasping is on the object to be grasped rather than the background information of the image or other targets. Therefore, we need to accurately determine the grasp region to reduce the area of grasp detection and improve the efficiency and robustness of grasp detection.

\section{Problem Formulation}\label{sec3}
\subsection{Grasp Representation}
Since Jiang et al.~\cite{10} proposed rotating rectangle boxes to represent the grasping pose, many researchers have built a grasp detection network based on the object detection network that can output multiple grasp rectangle boxes. The observation demonstrates that the gripper's width $h$ is a fixed parameter. Furthermore, the types of grippers chosen in each literature are distinct. As a result, we use a simplified representation of robot grasping similar to Morrison et al.~\cite{21}, and define the representation of robot grasping as:

\begin{equation}\label{equ2}
G=(p, \psi, w, q)
\end{equation}
where $p=(x, y, z)$ is the central position of the robot gripper in Cartesian coordinates, $\psi$ is the gripper's rotation angle around the z-axis, $w$ is the opening width of the gripper, $q$ is grasp confidence. Compared with the five-dimensional grasp representation, equation (2) can measure the probability of successful grasp and select the grasp with the highest quality value without evaluating from multiple grasp candidates. Assuming that the camera intrinsics parameters and calibration results are known, the robot derives the grasp pose in the plane from the depth image $I$ of size $H^{*} W$ :

\begin{equation}\label{equ3}
\wideparen{g}=\{\wideparen{p}, \wideparen{\psi}, \wideparen{w}, \wideparen{q}\}
\end{equation}
where $\wideparen{p}=(\wideparen{x}, \wideparen{y})$ is the pixel coordinate of the grasp center point, $\wideparen{\psi}$ is the rotation angle of the camera reference coordinate system around the z-axis, $\wideparen{w}$  is the opening width of the grasp gripper, and $\wideparen{g}$ is the grasp confidence. In order to perform grasping in image space on the robot, we convert $\wideparen{g}$ into grasp pose $g$ in the real world by the following formula:

\begin{equation}\label{equ4}
g=t_{r c}\left(t_{c i}(\wideparen{g})\right)
\end{equation}
where $t_{c i}$ is the transformation matrix from the image plane coordinate system to the camera coordinate system, and $t_{rc}$ is the transformation matrix from the camera coordinate system to the robot (world) coordinate system. We can do multiple grasps of images in the grasp dataset in image space, which can be denoted as:

\begin{equation}\label{equ5}
G=\{\Phi, W, Q\} \in \mathbb{R}^{3 \times W \times H}
\end{equation}
where $\Phi, W, Q$ are respective each $1 \times W \times H$, and contains the $\wideparen{\phi}, \wideparen{w}, \wideparen{q}$ values in each pixel.

\subsection{Gaussian-based guided training mehod}
\begin{figure}[!t]
	\centering
	\includegraphics[width=\linewidth]{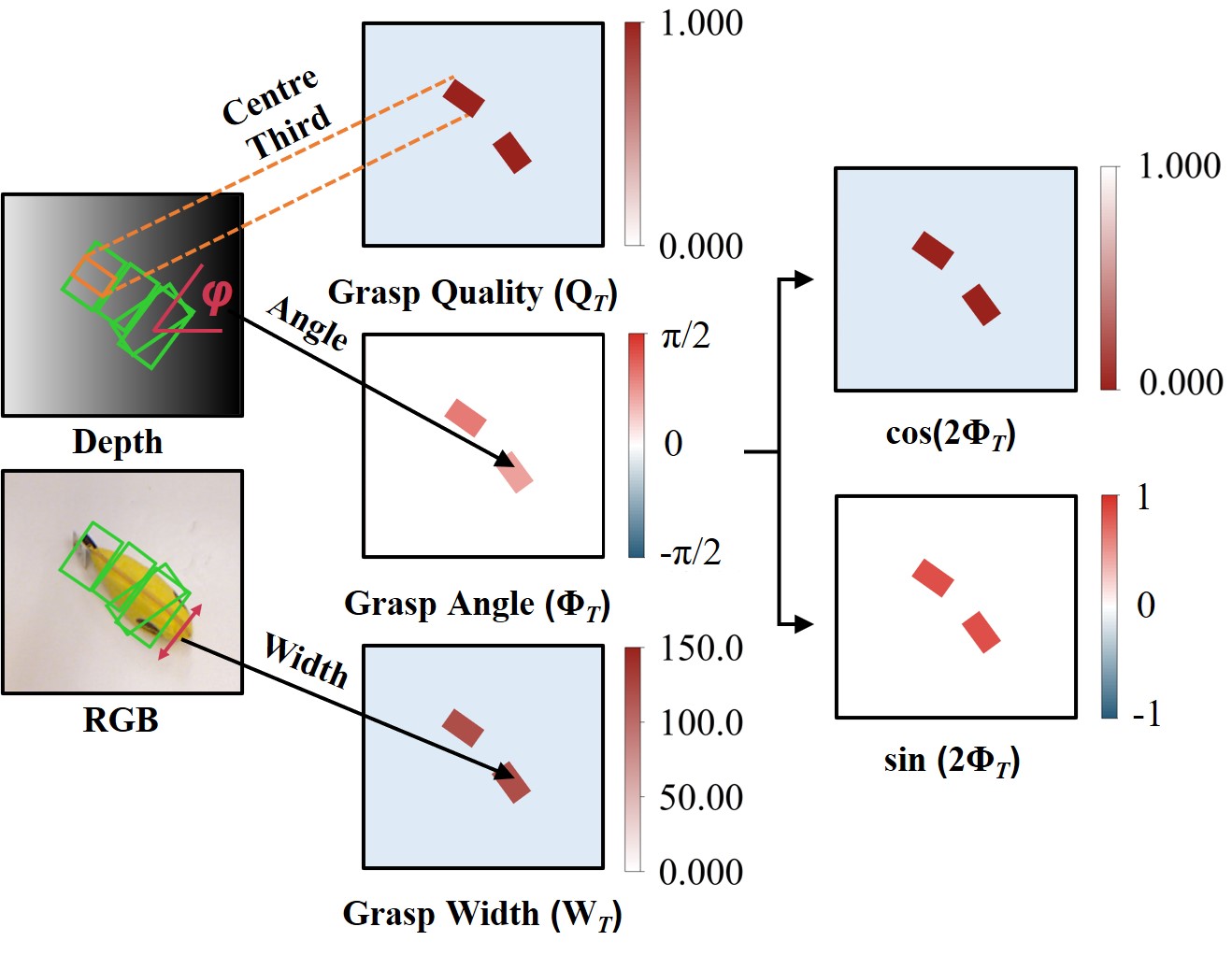}
	\caption{Taking Cornell Grasp Dataset as an example, the grasp representation and grasp maps of Morrison et al.~\cite{21}.}\label{fig3}
\end{figure}

Since the discrete rectangle box cannot cover all grasp positions on the object, reference~\cite{21} marks all pixels in the center 1/3 area of the grasp rectangle as grasp points with a grasp quality of 1, and the grasp point in the same area has the same grasp angle and width as the grasp rectangle, and then the pixel-level grasp pose is output by predicting the grasp quality, angle, and width for each pixel point, as shown in Fig.~\ref{fig3}.

Given that the grasp quality of the point closest to the object's central axis is frequently higher than that of the point closest to the object's edge, in~\cite{24}, the grasp quality of the points in the rectangle is expressed as a two-dimensional Gaussian distribution with the mean value located at the center of the grasp rectangle box, the grasp quality of the center point is specified as 1, and the grasp quality of the surrounding points decreases, emphasizing the central point's importance. However, the point with the highest grasp quality on the object is not just the center point of the rectangle, but should be all points located on the central axis. Aiming at the above problems, this paper uses an improved one-dimensional Gaussian distribution to represent the grasp quality of points in the grasp rectangle and introduces a Gaussian distribution to represent the quality of the grasp angle. For any ground truth in the Cornell Grasp Dataset, we generate a Gaussian-based guided training method as follows:

\subsubsection{The quality of grasp points}
\begin{figure}[!t]
	\centering
	\includegraphics[width=\linewidth]{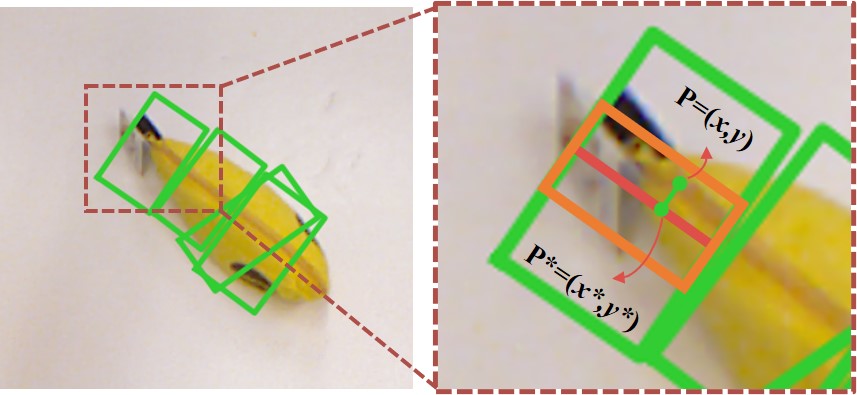}
	\caption{Visualization of grasping point representation based on Gaussian guidance.}\label{fig4}
\end{figure}

We intercept the middle 1/3 of the grasp rectangle $G$ and express the grasp quality of points in the region as a Gaussian distribution with the mean value located on the central axis. The minimum grasp quality is 0.5. For any point  $p=(x,y)$, the grasp quality $q$ of this point is defined as follows:

\begin{equation}\label{equ6}
q=\exp \left(-\frac{L\left(p, p^{*}\right)}{2 \sigma_{q}^{2}}\right)
\end{equation}

\begin{equation}\label{equ7}
L\left(p, p^{*}\right)=\left(x-x^{*}\right)^{2}-\left(y-y^{*}\right)^{2}
\end{equation}
where $p^{*}=\left(x^{*}, y^{*}\right)$ is the intersection point of $p$ perpendicular to the central axis, as shown in Fig.~\ref{fig4}. Except for all points after the middle 1/3 region, the grasp quality is marked as 0.

\subsubsection{The quality of grasp angles}
\begin{figure}[!t]
	\centering
	\includegraphics[width=5cm]{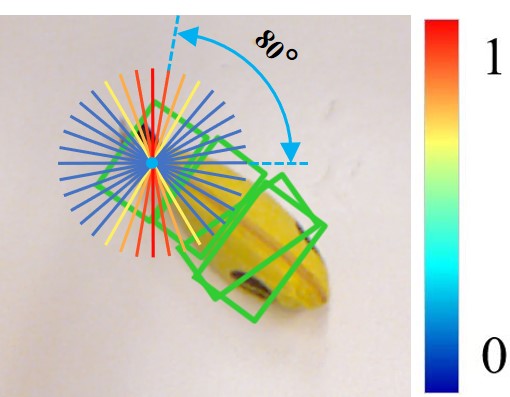}
	\caption{Visualization of grasping Angle quality when grasp angle $\theta$ = $80^{\circ}$.}\label{fig5}
\end{figure}

For each pixel in the image, we predict the grasp angle by classification. We define the grasp angle as a radian value in the $[0, \pi]$ interval, and obtain the category corresponding to the labeled grasp angle $\theta$ according to the following formula:

\begin{equation}\label{equ8}
k=\left[\frac{\theta}{\pi} K\right]
\end{equation}
where $K$ is equal to 18. For classification prediction, we define $\Theta$ as a unit vector with a dimension of $1 \times K$, where the value $\Theta_{i}$ at the $i$ position represents the grasping quality of the angle value $\frac{i}{K} \pi$, and $\Theta_{k}=1$, indicating that the labeled grasping angle $\theta$ has the highest quality. Given that angles close to the grasping angle $\theta$  still have high grasp quality, we use a Gaussian distribution with mean $\theta$ to determine the value of $\Theta$ in the vector. For the value $\Theta$ at the $i$-th position of vector $\Theta_{i}$, it is defined as follows:

\begin{equation}\label{equ9}
\Theta_{i}=\exp \left(-\frac{(i-k)^{2}}{2 \sigma_{a}^{2}}\right),|i-k| \leq t h
\end{equation}
where $th$ is the error range of the grasp angle. Considering that the object can still be grasped when the closing angle of the manipulator is $30^{\circ}$ different from the marked grasp angle, we set $t h$ to 3 . When $|i-k|>t h, \Theta_{i}=0 .$ As shown in Fig.~\ref{fig5}.

\section{Proposed Method}
\begin{figure*}[htb]
	\centering
	\includegraphics[width=\linewidth]{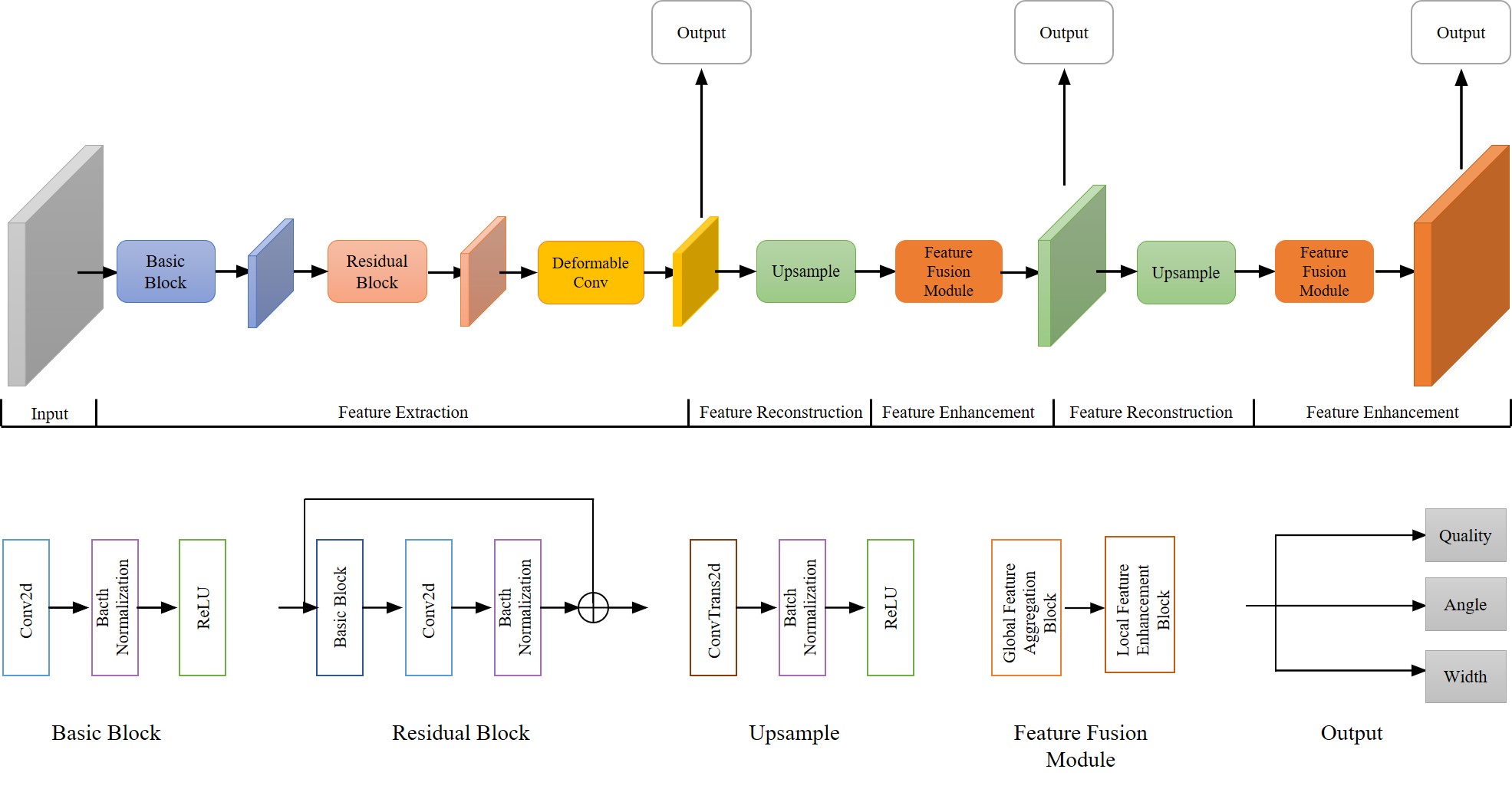}
	\caption{The architecture of proposed new generative grasp detection network.}\label{fig6}
\end{figure*}

In this section, we propose a novel network architecture for grasp detection, as shown in Fig.~\ref{fig6}. o obtain more information about the image, the input is first passed through a convolutional layer for feature extraction and downsampling. To learn the shape information of different objects, we use deformable convolution to replace the last convolutional layer in the feature extractor. Simultaneously, we propose a global-local attention  network for feature fusion in the feature enhancement stage after feature reconstruction. The convolution layer is used to obtain global feature information through the global feature aggregation block (GFAB), and global average pooling and global feature aggregation are used to obtain global feature information. The final enhanced feature map is obtained by using convolutional layers and dense connections in the local feature enhancement block (LFEB). In the output prediction stage, we predict the grasp values of the  feature maps of three different scales, obtain the quality, angle, and width under different scales, and calculate the multi-scale loss with the ground truth to improve the ability of the network to grasp detection objects under different scales. Each module is described in detail below.

\subsection{Network architecture}
Our proposed network is composed mainly of four parts: feature extraction, feature enhancement, feature reconstruction, and output prediction. Feature extraction consists of $3 \times 3$ convolutional layers, with a total of four times downsampling. After downsampling, five residual blocks are used to further extract features. Generally, the shape of the object is not fixed, and effective modeling of the object improves grasping performance. As a result, in the last stage of feature extraction, deformable convolution is used to obtain the object's shape information and increase the deformation modeling ability.  A 2D convolution process with a convolution kernel size of $3 \times 3$ is often divided into two parts. Each point on the feature map $F_{i}$ is first sampled in a grid $R$, and then these sampling points are weighted by the weight $w$. $R$ is the convolution kernel's receptive field, and $R$ of $3 \times 3$ convolution kernel is represented as follows:

\begin{equation}\label{equ10}
R=\{(-1,-1),(-1,0), \ldots,(0,1),(1,1)\} 
\end{equation}

For point $P_{0}$ on the feature map $F_{i}$ extracted from each grasped image, the equation can be represented as follows:

\begin{equation}\label{equ11}
F_{o}\left(p_{0}\right)=\sum_{p_{n} \in R} w\left(p_{n}\right) \cdot F_{i}\left(p_{0}+p_{n}\right)
\end{equation}
where $p_{n}$ is sampled position. In deformable convolution, the grid $R$ is accompanied by different offsets
$\left\{\Delta p_{n} \mid n=1, \ldots, N\right\}$, where $N$ is the number of points in $R$.  So  Equation~\eqref{equ12} can be defined as:

\begin{equation} \label{equ12}
F_{o}\left(p_{0}\right)=\sum_{p_{n} \in R} w\left(p_{n}\right) \cdot F_{i}\left(p_{0}+p_{n}+\Delta p_{n}\right)
\end{equation}

We perform bilinear interpolation sampling on each sampling point to obtain the final output feature map $F_{o}$ to learn the shape features of the object.

In the feature enhancement stage, we perform global-local feature fusion on the feature map to obtain finer object information and attenuate unwanted noise. The module consists of GFAB and LFEB, the details were described in Section B. Each feature enhancement module is followed by a feature reconstruction module that uses transposed convolution to restore the image's size. Finally, we make an output prediction and calculate the loss of multiple scales using different scales.

\subsection{Global-local feature fusion module}
Following the feature extraction stage, the image's rich feature information frequently contains some redundant noise. A lot of attention mechanism models have been proposed in the field of computer vision to solve the problem of feature redundancy and let the model focus on more interesting parts of the image. To ensure that the model is focused on the object rather than on the background noise during grasping detection, we design a global-local feature fusion module during the feature enhancement stage to extract more useful information from the grasped image. It is composed primarily of the GFAB (Global feature aggregation block) and the LFEB (local feature enhancement block).

\begin{figure}[htbp]
	\centering
	\includegraphics[width=\linewidth]{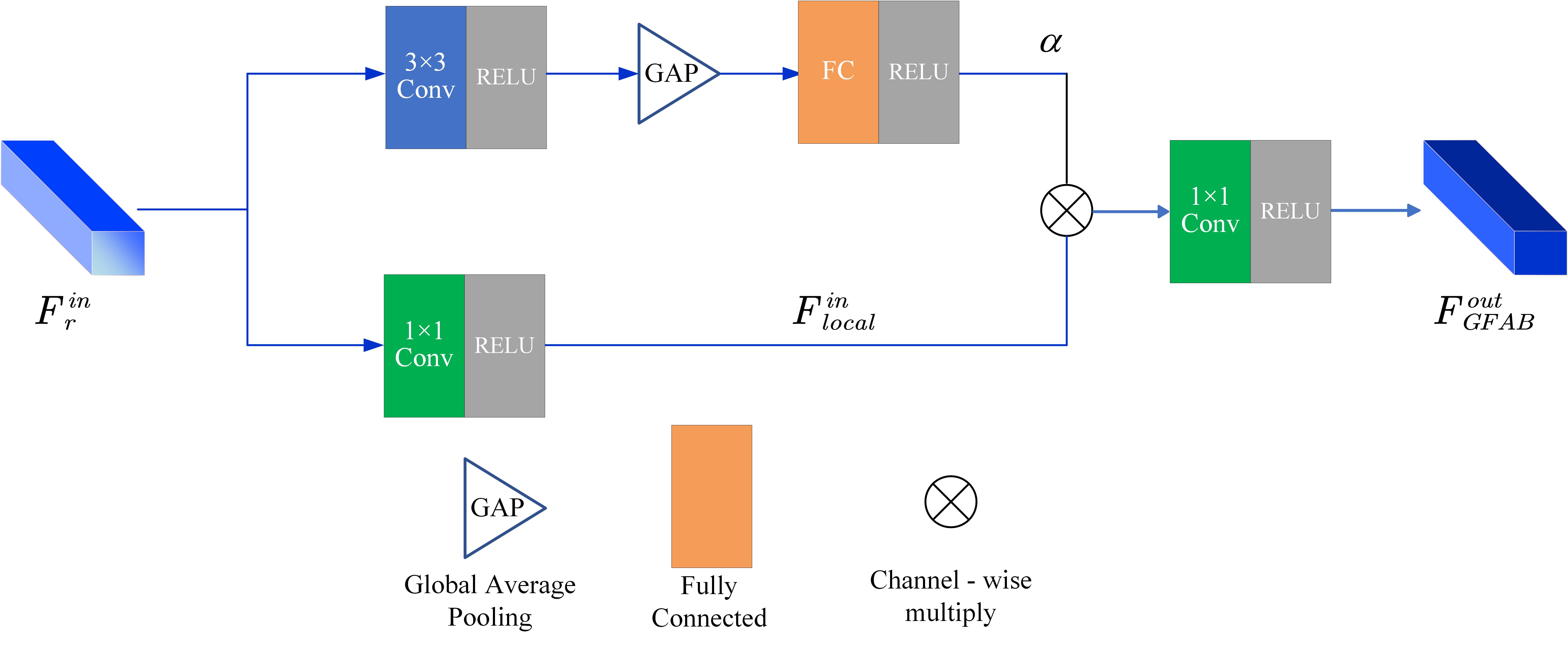}
	\caption{The architecture of proposed global feature aggregation block.}\label{fig7}
\end{figure}

\textbf{Global Feature Aggregation Block.} The GFAB module is divided into two branches: global channel weighting and local feature extraction. For the input feature map $F_{r}^{in}$, after passing through the convolutional layer through the channel weighting branch and relu activation layer, the global average pooling and fully connected layer are used to aggregate the information of global features to obtain the weight $\alpha$ on the channel; through the $1 \times 1$ convolution layer and relu activation layer of the local feature extraction branch, feature map $F_{r}^{in }$ obtains a local feature map $F_{local }^{in}$, and then products with $\alpha$ on the channel to obtain the output $F_{G F A B}^{out }$ of GFAB via the final $1 \times 1$ convolution and relu layer.. It can be defined as:

\begin{equation}\label{equ13}
F_{G F A B}^{o u t}=C R\left(\alpha \times F_{local}^{in }\right)
\end{equation}
where $CR$ represents $1 \times 1$ convolution layer and RELU activation layer. As shown in Fig.~\ref{fig7}.

\begin{figure}[htbp]
	\centering
	\includegraphics[width=\linewidth]{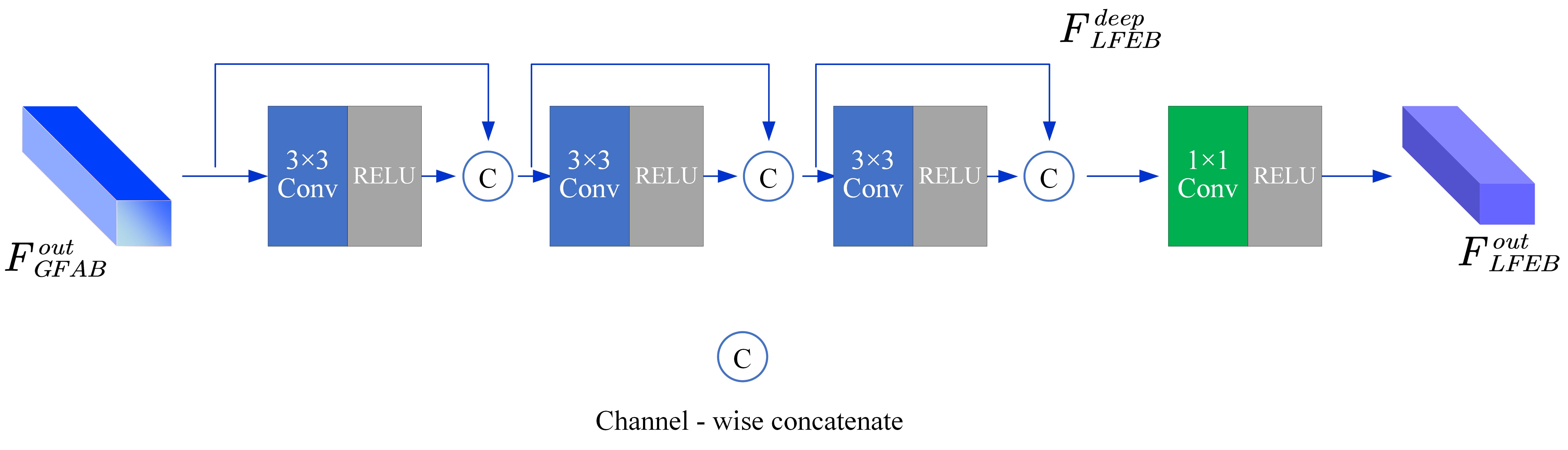}
	\caption{The architecture of proposed local feature enhancement block.}\label{fig8}
\end{figure}

\textbf{Local feature enhancement block.} The LFEB module is used to obtain finer features after GFAB enhancement. For the output $F_{G F A B}^{out}$ of GFAB module, we use a densely connected convolutional layer to obtain the features $F_{L F E B}^{deep}$ that fuse different stages, obtain the final fusion feature $F_{L F E B}^{out}$, and finally use $1 \times 1$ convolution to obtain the output result. The visualization is shown in Fig.~\ref{fig8}. It can be defined as:

\begin{equation}\label{equ14}
F_{L F E B}^{out}=C R\left(F_{L F E B}^{deep}\right)
\end{equation}
where $CR$ represents $1 \times 1$ convolution layer and RELU activation layer.

\subsection{Loss function}
For the input image $I$, the grasp label can be represented as $Y=\left\{y_{1}, \ldots, y_{n}\right\}$. After the grasp network proposed we can get the corresponding output is $\hat{Y}=\left\{\hat{y}_{1}, \ldots, \hat{y}_{n}\right\}$. In this paper, we use a multi-scale loss to predict the corresponding output $\hat{Y^s}$ at three stages of feature reconstruction, where $s = 1, 2$, and $3$ , corresponding to 4 times downsampling, 2 times downsampling, and original size, respectively. In each scale, we calculate smooth L1 loss with the label of the corresponding scale. For a certain scale $s$, the loss function can be defined as:

\begin{equation}\label{equ15}
L^{s}\left(\hat{Y^{s}}, Y^{s}\right)=\sum_{i}^{n} \l_{1}\left(\hat{y}_{i}^{s}-y_{i}^{s}\right)
\end{equation}
where $n$ is the number of grasp positions and each prediction has three components of quality, angle, and width. $l_{1}$ represents smooth L1 loss, it can be defined as:

\begin{equation}\label{equ16}
l_{1}(x)=\left\{\begin{array}{cc}
(\sigma x)^{2} / 2, & \text { if }|x|<1 \\
|x|-0.5 / \sigma^{2}, & \text { otherwise }
\end{array}\right.
\end{equation}

In smooth L1 loss, $\sigma $ represents for the hyperparameter, which is used to adjust the smooth index. Total loss $\mathcal{L}$ can be formulated as:
\begin{equation}\label{equ17}
\mathcal{L}=\frac{1}{s}\sum\limits_{i=1}^{s}{L^{i}}.
\end{equation}
where s = 3, which represents three different scales.

\section{Experiments and Result} \label{sec5}
\subsection{Datasets}
The Cornell Grasping Dataset and the Jacquard Dataset are widely used as verification standards for robotic grasp detection. As a result, we train our method on these two datasets and compare it to other algorithms.

\textbf{Cornell Grasp Dataset.} Cornell Grasp Dataset contains 240 different objects, a total of 885 images and point cloud data in the global coordinate system. The image and point cloud data are aligned. Each image is labeled with multiple good ground truth of grasp points, with a total of 5110 positive and 2909 negative grasp rectangles. As with many previous studies, we divided the datasets into two distinct categories:

1) Image-wise split: The datasets are randomly divided into training sets and test sets. This is mainly to test the adaptability of the network model in detecting the same object at different positions and angles.

2) Object-wise split: The datasets are splited according to object instances, and the generalization of the model to unseen objects is tested by using data that did not appear in the training set before. 

In addition, compared with other deep learning datasets, the Cornell dataset is very small. Therefore, in order to avoid overfitting of the model, we expand the data set by cropping, random flipping, and translation before training the network.

\textbf{Jacquard Dataset.} The Jacquard dataset is a simulation grasp dataset nearly 50 times larger than the Cornell dataset, including 54485 images of 11619 objects. Simultaneously, it has multiple data types, including rgb-d and mask, which does not need data enhancement like the Cornell dataset. The dataset is split into training and test sets in a 5:1 ratio.

\subsection{Implementation details}
Our method is implemented in Pytorch 1.5.0, and the experimental platform is based on a single NVIDIA GeForce RTX 2080Ti (Pascal architecture with 12G memory) and the Ubuntu 16.04 operating system. During training, we used the Adam optimizer to propagate back. A learning rate of 0.001 at the start and a total training cycle of 60. When the loss is no longer reduced in ten consecutive batches, the initial learning rate is reduced to a tenth of its original value.

To conduct an objective evaluation of our work, we use the same evaluation index as many previous studies, see equation~\eqref{equ1}. Due to the limitations of the rectangle metric and in order to reduce the false-positive grasps, we will improve the standard, that is, increase the Jaccard threshold to 0.30, 0.35, and 0.4 and reduce the angle difference between predicted and ground truth, and then test the performance on two public datasets. In this paper, the Jaccard index is only used to evaluate the model, not to the matching strategy. equation~\eqref{equ1}. 

\subsection{Result and analyses}
To verify the performance of our proposed method, we compare it with the previous algorithms on Cornell Grasp Dataset, and the results are shown in Table 1. Our method achieves 99.0$\%$ and 98.3$\%$ in image-wise and object-wise, respectively, and achieves the state-of-the-art performance. In addition, the inference time of our method is only 10 ms, which indicates real-time performance. Similarly, we also carry out experimental verification on Jacquard Dataset, as shown in Table 2. Additionally, our method also achieves the excellent performance, reaching 95.9$\%$.

\begin{table}[!t]
	\centering
	\caption{Performance evaluation of different methods in the Cornell Grasp Dataset.}\label{tab1}
	\setlength\tabcolsep{2.5pt}
	\begin{tabular}{@{}lcccc@{}}
		\toprule
		\multirow{2}{*}{\textbf{Method}} & \multirow{2}{*}{\textbf{Input   Modality}} & \multicolumn{2}{c}{\textbf{Detection Accuracy}} & \multirow{2}{*}{\textbf{Inference Time}} \\ \cmidrule(lr){3-4}
		&                                            & \textbf{IW}            & \textbf{OW}            &                                          \\ \midrule
		Jiang et al.~\cite{10}            & RGB-D                                      & 60.5\%                 & 58.3\%                 & 5000ms                                   \\
		Lenz et al.~\cite{9}              & RGB                                        & 73.9\%                 & 75.6\%                 & 1350ms                                   \\
		Redmon et al.~\cite{17}         & RGB-D                                      & 88\%                   & 87.7\%                 & 76ms                                     \\
		Morrison et al.~\cite{21}       & D                                          & 73.0\%                 & 69.0\%                 & 19ms                                     \\
		Guo et al.~\cite{18}            & RGB-D                                      & 93.2\%                 & 89.1\%                 & -                                        \\
		Kumra et al.~\cite{25}            & RGB                                        & 89.2\%                 & 88.9\%                 & 103ms                                    \\
		Chu et al.~\cite{19}              & RGB-D                                      & 96.0\%                 & 96.1\%                 & 96ms                                     \\
		Zhou et al.~\cite{20}           & RGB                                        & 97.7\%                 & 96.6\%                 & 117ms                                    \\
		Wang et al.~\cite{26}           & RGD                                        & 94.4\%                 & 91.0\%                 & 8ms                                      \\
		Song et al.~\cite{27}           & RGB                                        & 96.2\%                 & 95.6\%                 & -                                        \\
		Kumra et al.~\cite{22}           & RGB-D                                      & 97.7\%                 & 95.9\%                 & 20ms                                     \\
		Chen et al.~\cite{12}           & RGB                                        & 93.5\%                 & 92.2\%                 & 146ms                                    \\
		Ours                             & RGB-D                                      & 99.0\%                 & 98.3\%                 & 9ms                                      \\ \bottomrule
	\end{tabular}
\end{table}
\begin{table}[!t]
	\centering
	\caption{Performance evaluation of different methods in the Jacquard Dataset.} \label{tab2}
	\begin{tabular}{@{}lcc@{}}
		\toprule
		\textbf{Method}          & \textbf{Input Modality} & \textbf{Detection Accuracy} \\ \midrule
		Morrison et al.~\cite{21}& D                       & 84\%                        \\
		Song et al.~\cite{27}    & RGB                     & 93.6\%                      \\
		Zhou et al.~\cite{20}     & RGB                     & 91.8\%                      \\
		Kumra et al.~\cite{22}    & RGB-D                   & 94.6\%                      \\
		Cao et al.~\cite{24}      & RGB-D                   & 95.6\%                      \\
		Chen et al.~\cite{12}     & RGB-D                   & 91.2\%                      \\
		Ours                     & RGB-D                   & 95.9\%                      \\ \bottomrule
	\end{tabular}
\end{table}

\begin{figure}[htbp]
	\centering
	\includegraphics[width=8.5cm]{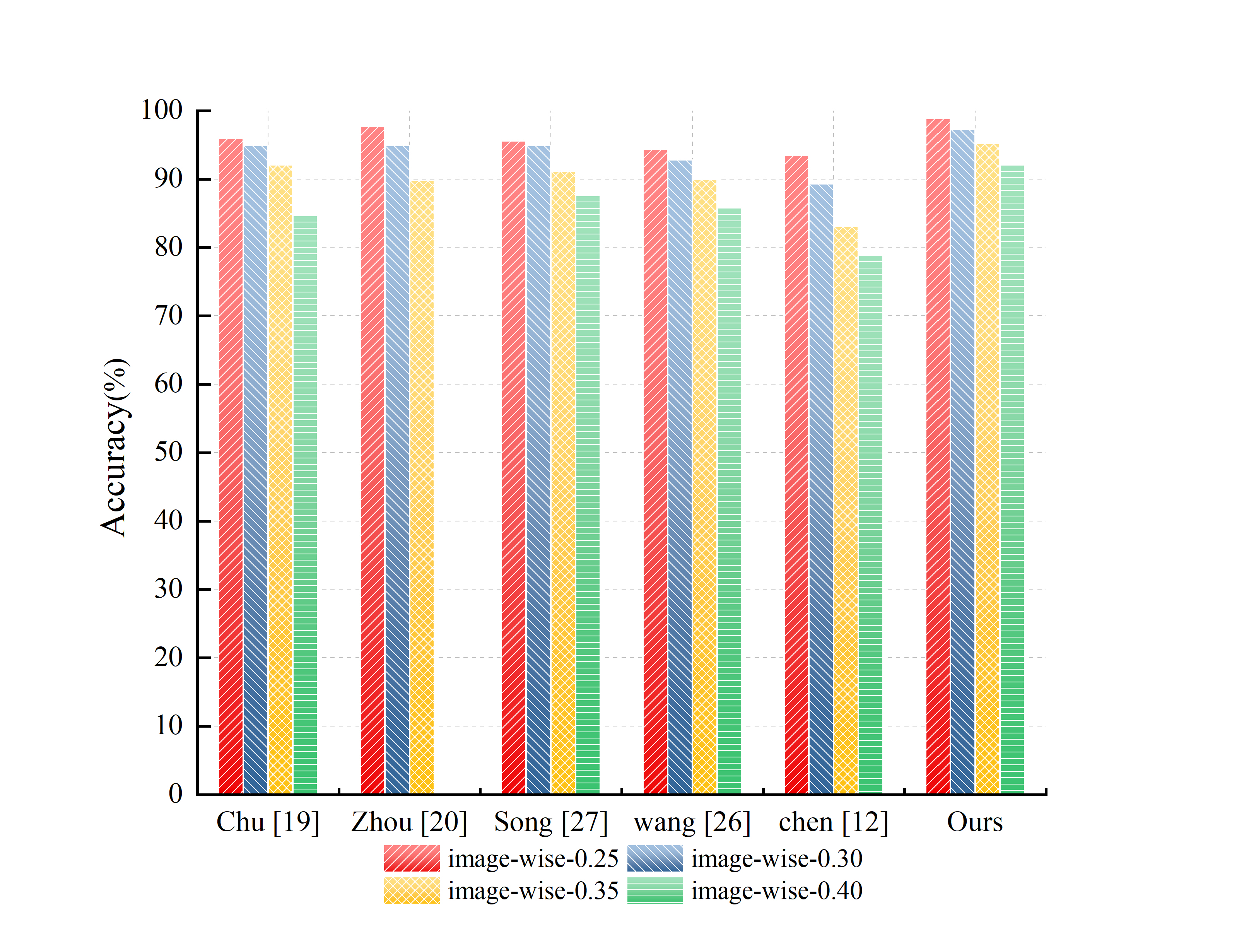}\\
	\includegraphics[width=8.5cm]{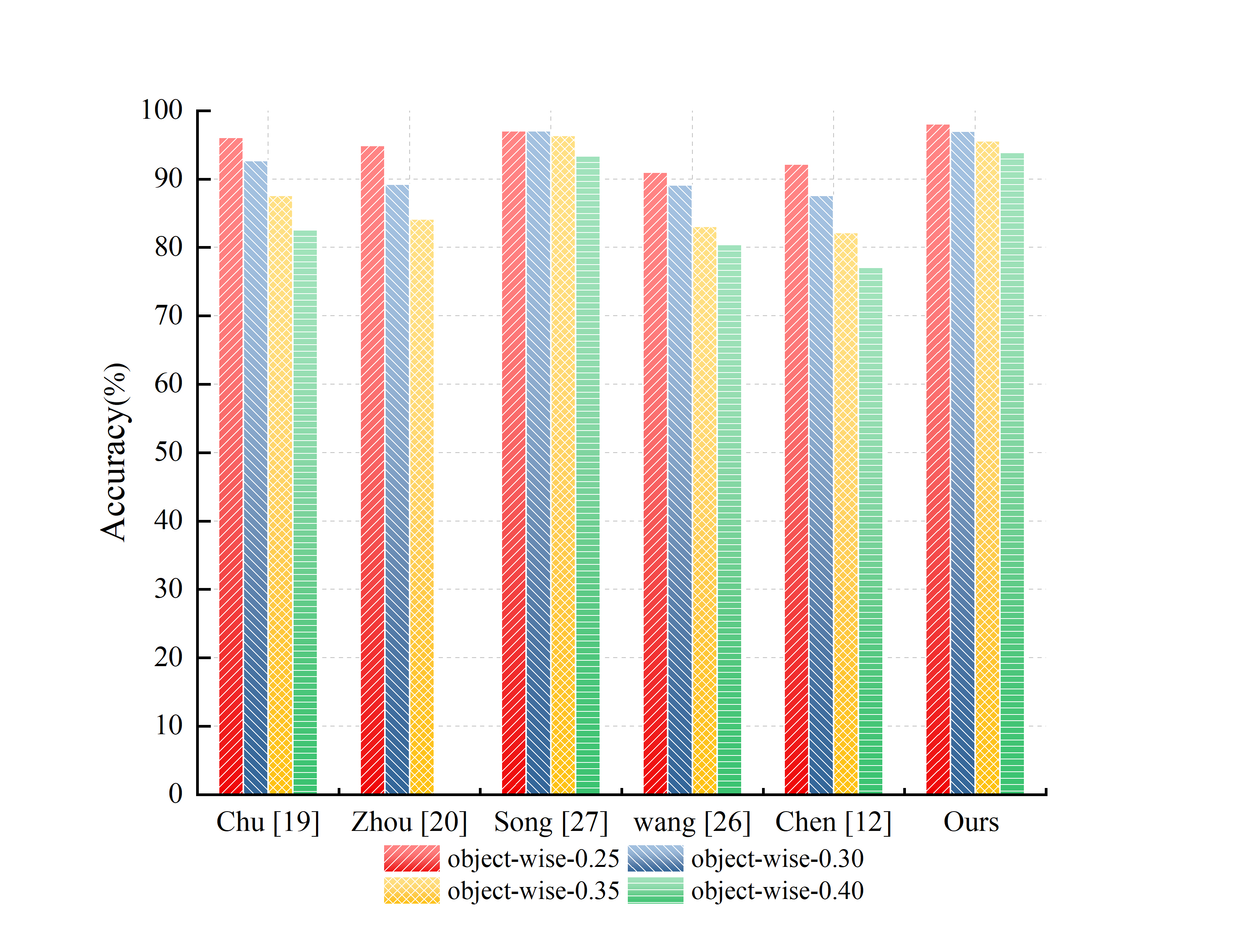}
	\caption{Experimental results on Cornell Grasp Dataset under different Jaccard indexes with various methods.}\label{fig9}
\end{figure}

To further verify the suppression of false-positive grasps by our method, we increase the jaccard index in the rectangle metric to 0.3, 0.35, 0.4, 0.45, and the angle difference between the prediction and the ground truth decreases in order. Take Cornell as an example, compared it with several excellent grasping methods, as shown in the Fig.~\ref{fig9}. It's worth noticing that while the rectangle metric has increased, the performance of other ways has deteriorated to varied degrees, but our method still retains a 93.9$\%$ accuracy at jaccard $>$ 0.4, which is far better than all other methods, and the benefit becomes more and more obvious with the increase of the jaccard value. Similarly, we gradually reduce the angle difference between the predicted and ground truth grasp angles to $10^{\circ}$ for testing. Our method is the superior at each angle difference, which further verify the necessity and effectiveness of our method.

\begin{figure}[htbp]
	\centering
	\includegraphics[width=8cm]{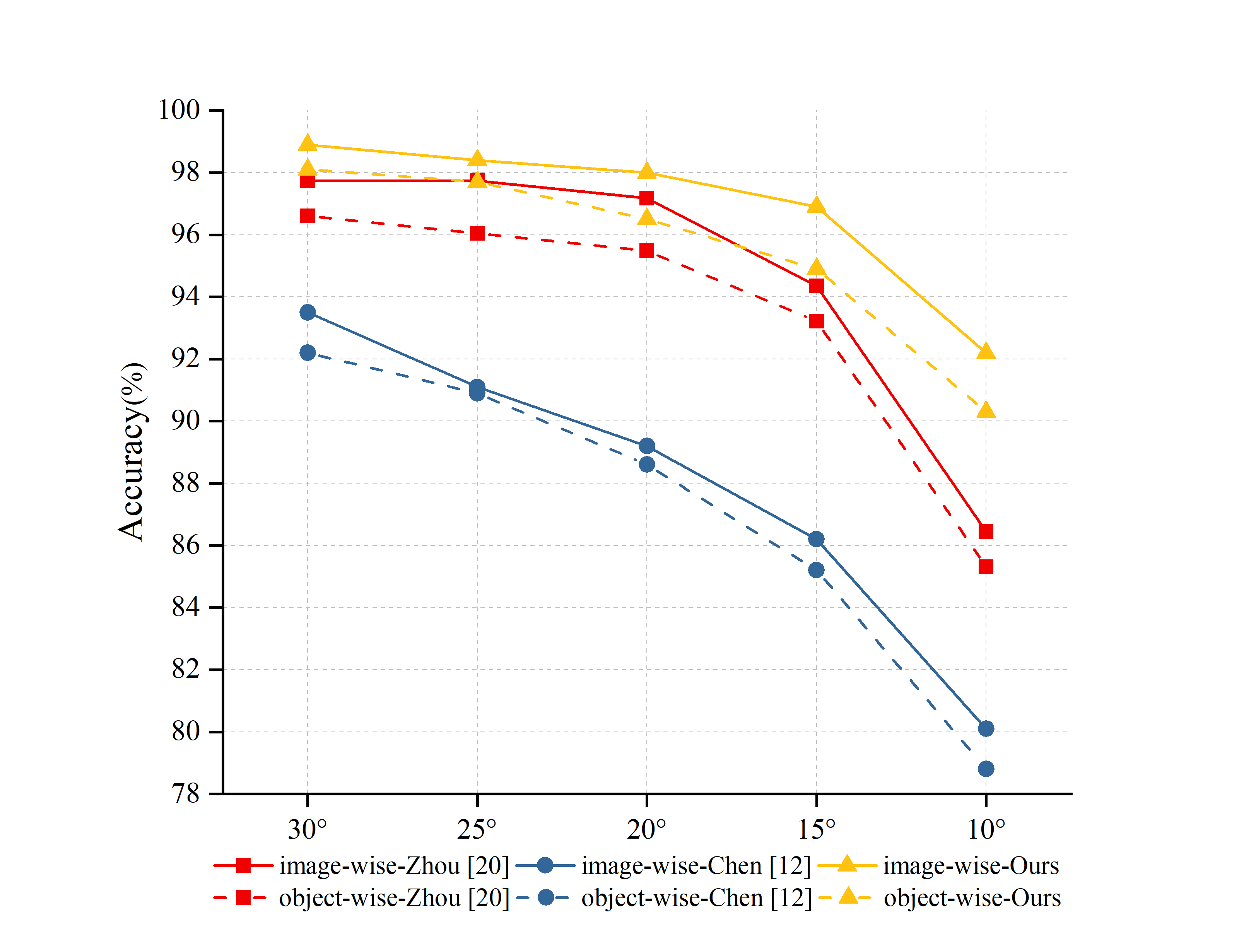}
	\caption{Experimental results on Cornell Grasp Dataset under different angle with various methods.}\label{fig10}
\end{figure}

\begin{figure}[!t]
	\centering
	\includegraphics[width=\linewidth]{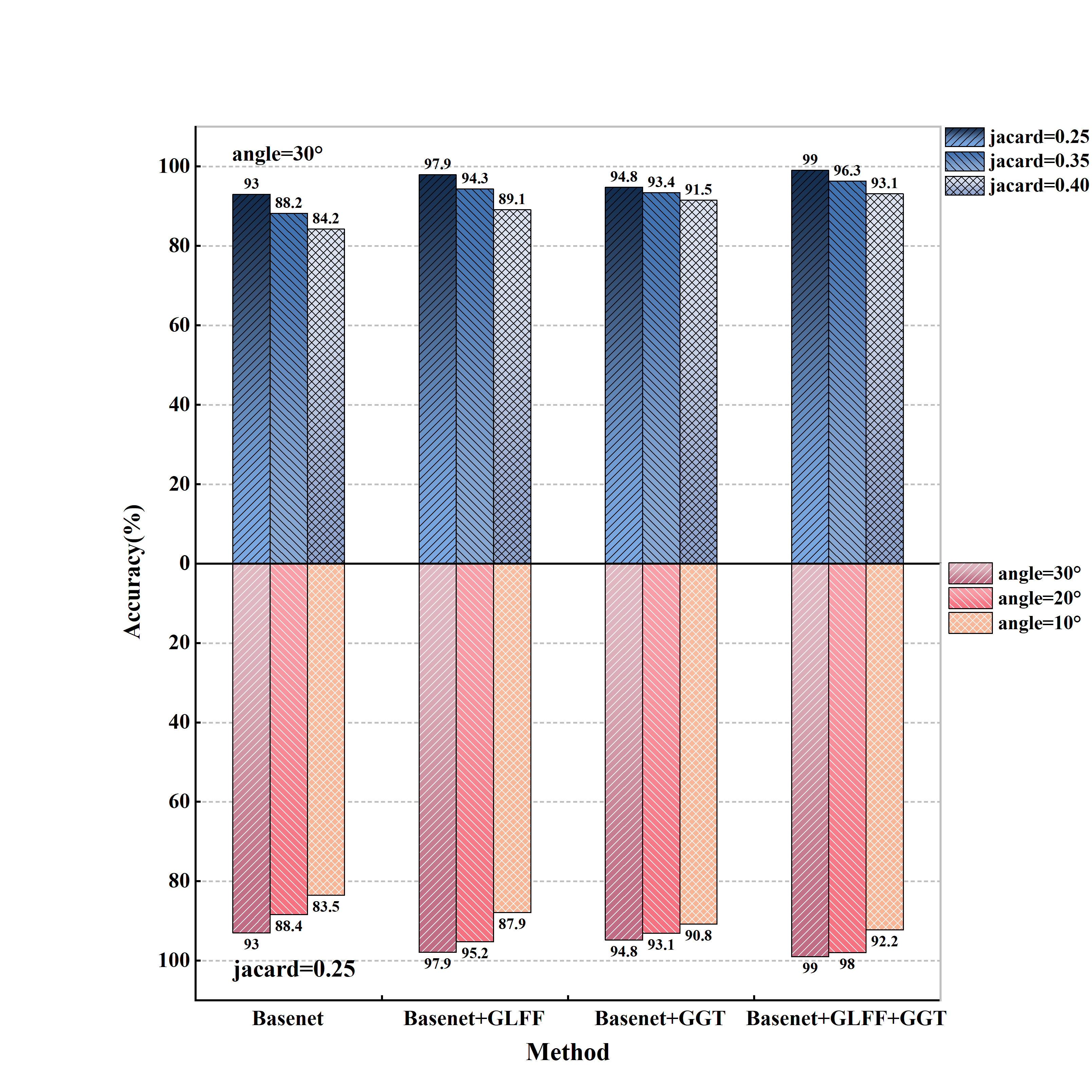}
	\caption{The effect of different modules on network performance.}\label{fig11}
\end{figure}

We also conduct ablation experiments to discuss how different modules can improve the overall architecture. Taking Cornell as an example, when the rectangle metric is initialized, the performance improvement of the GLFF (global-local feature fusion) module is much greater than that of the GGT (gaussian-based guided training) module. As the rectangle metric increases, the performance of the GLFF module decreases significantly, while the performance of the GGT module decreases slowly. This is because the GLFF module focuses on increasing the network's accuracy, and the GGT module on reducing false positive grasps. The best performance can be achieved when all modules are combined. The visualization is shown in Fig.~\ref{fig11}.

\begin{figure}[!t]
	\centering
	\includegraphics[width=\linewidth]{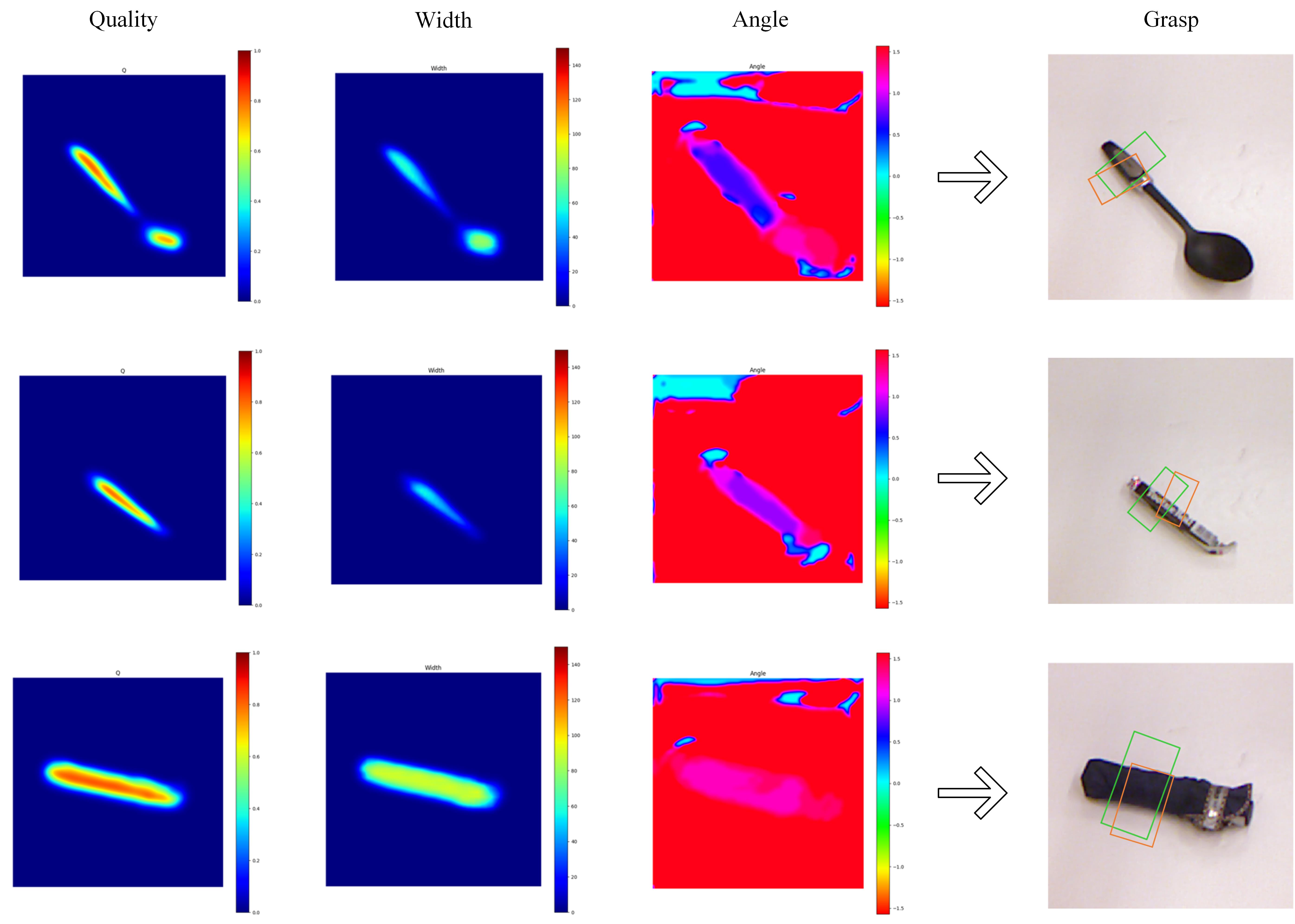}
	\caption{Examples of detection results on Cornell. The columns represent the quantity, width, and angle output, respectively. The last column's green grasp rectangle is ours, while the yellow is~\cite{12}.}\label{fig12}
\end{figure}

Fig.~\ref{fig12} shows the visualization of the correction of false-positive grasps by our method. Taking Chen et al.~\cite{12} as an example, although they have demonstrated excellent performance on the Cornell Grasp Dataset, there are still some false-positive grasps. Our method clearly corrects these false-positive grasps, and the prediction value is more consistent with human gripping behavior, which increases the success rate and stability of the real robot grasping experiment.

\subsection{Real-world robotic grasping}
\begin{figure}[htbp]
	\centering
	\includegraphics[width=\linewidth]{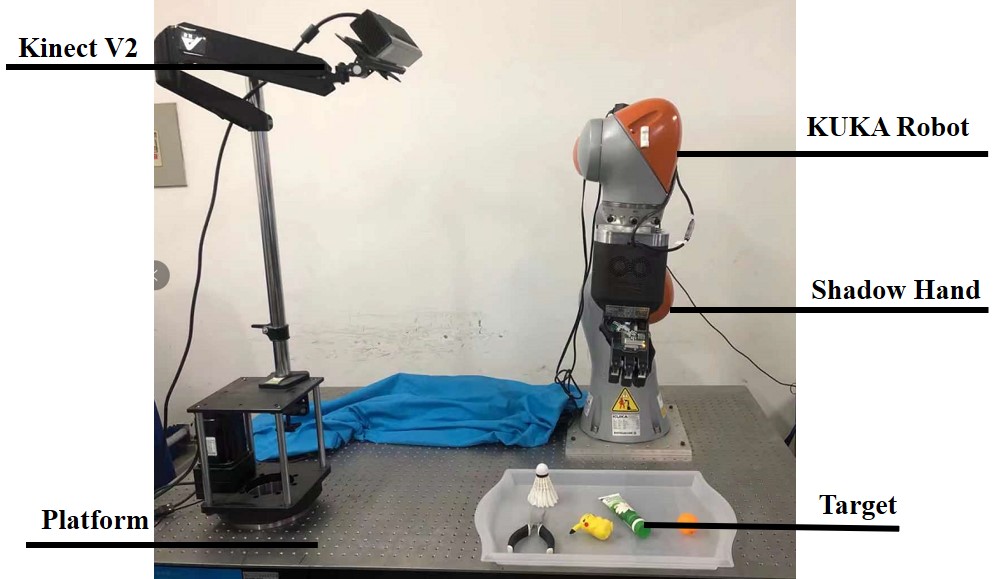}
	\caption{Overview of the built robotic grasping system.}\label{fig13}
\end{figure}

To verify whether the method proposed in this paper can improve the success rate of real robot grasping, we build a reasonable experimental system for verification, which is composed of KUKA (LBR iiwa 14 r820), a shadow hand, and a platform. A Kinect v2 real-time camera was used to acquire image data. To obtain a more precise measurement without encroaching on the robot's space, we mounted the camera on the robot's left front and slightly above its line of sight to ensure global information throughout the experiment. The robot experimental system is shown in Fig.~\ref{fig13}.

\begin{figure}[!t]
	\centering
	\includegraphics[width=\linewidth]{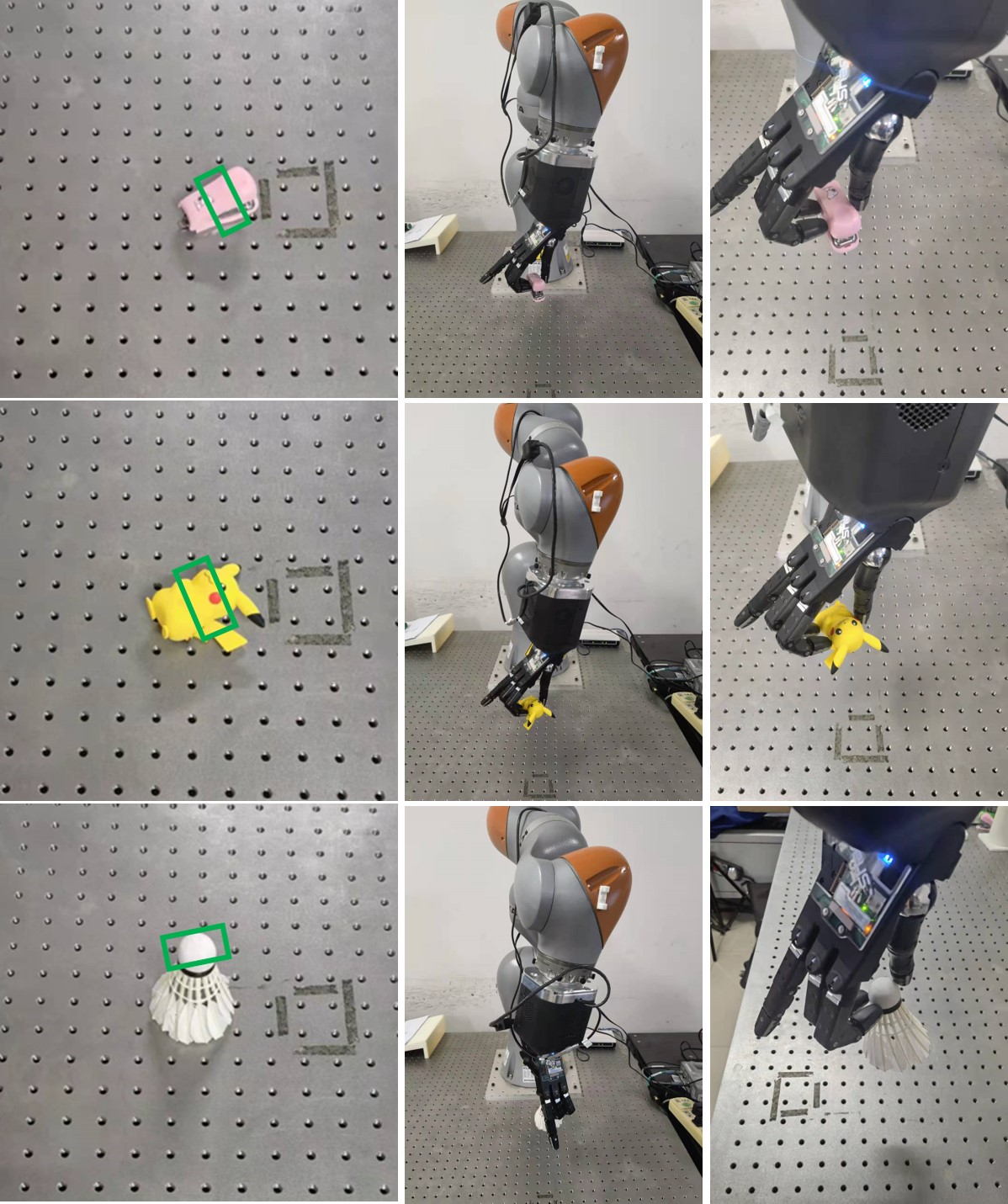}
	\caption{Visualization of real-world grasp detection. The first column is the grasp rectangles. The second and third columns are the results of the panoramic grasp from different angles.}\label{fig14}
\end{figure}

We still use~\cite{12} to compare with our method. In the grasping experiment, more than 10 objects were chosen, all of which did not exist in the training data set. Objects are randomly put in different places and directions during the experiment, and each object was grasped 80 times. Our method has a success rate of 94$\%$ on average, which is nearly equal to the performance in the dataset. While~\cite{12} has a success rate of only 72$\%$, it is significantly lower than the test results in the dataset. It is because the false-positive grasps does not conform to the grasping conditions of the real robot, resulting in grasping failure or process instability, which results in the object falling. This further verifies the effectiveness and robustness of our method. Fig.~\ref{fig14} shows some visualization results.

\section{Conclusion}
In this paper, we present a novel generative convolutional neural network model to increase the accuracy and robustness of real-world robot grasping detection tasks. To begin with, a Gaussian distribution is employed, which standardizes the position and angle information of the grasping rectangle to the maximum extent. On this basis, a deformable convolution and a global local feature fusion method are presented to guide the network's attention to the grasped object's features. Our method outperforms other methods on the Cornell and Jacquad datasets. Finally, we perform a real-world scenario experiment to prove the efficacy of the proposed method.

\bibliographystyle{IEEEtran}
\bibliography{mybib}

%
%
%
%
%

\vfill

\end{document}